# Adversarial-Ensemble Kolmogorov Arnold Networks for Enhancing Indoor Wi-Fi Positioning: A Defensive Approach Against Spoofing and Signal Manipulation Attacks

Mitul Goswami[1], Romit Chatterjee[1], Somnath Mahato[2], and Prasant Kumar Pattnaik[1]

*School of Computer Engineering, Kalinga Institute of Industrial Technology, Patia, Bhubaneswar, 751024, India[1], Meteorological Training Institute, India Meteorological Department, Pashan, Pune, 411008, India[2]*

***Abstract*—The research presents a study on enhancing the robustness of Wi-Fi-based indoor positioning systems against adversarial attacks. The goal is to improve the positioning accuracy and resilience of these systems under two attack scenarios: Wi-Fi Spoofing and Signal Strength Manipulation. Three models are developed and evaluated: a baseline model ($M_{Base}$), an adversarially trained robust model ($M_{Rob}$), and an ensemble model ($M_{Ens}$). All models utilize a Kolmogorov-Arnold Network (KAN) architecture. The robust model is trained with adversarially perturbed data, while the ensemble model combines predictions from both the base and robust models. Experimental results show that the robust model reduces positioning error by approximately 10% compared to the baseline, achieving 2.03 meters error under Wi-Fi spoofing and 2.00 meters under signal strength manipulation. The ensemble model further outperforms with errors of 2.01 meters and 1.975 meters for the respective attack types. This analysis highlights the effectiveness of adversarial training techniques in mitigating attack impacts. The findings underscore the importance of considering adversarial scenarios in developing indoor positioning systems, as improved resilience can significantly enhance the accuracy and reliability of such systems in mission-critical environments.**

***Index Terms*—Indoor Positioning System, Wi-Fi Spoofing, Signal Strength Manipulation, Kolmogorov-Arnold Network, Adversarial Attacks, Ensemble Model**

## I. INTRODUCTION

Wi-Fi-based Indoor Positioning Systems (IPS) utilize existing Wi-Fi infrastructure to determine the location of devices within indoor environments. By analyzing Received Signal Strength Indicator (RSSI) values from multiple access points, these systems can achieve precise location estimates. R. Parker et al. used road maps, vehicle kinematics, and intervehicle distance measurements based on the Received-Signal-Strength Indicator to determine the relative placements of automobiles in a cluster [1]. The growing importance of Wi-Fi-based IPS is evident across various applications, including smart buildings for efficient space utilization, navigation in complex indoor settings like airports and malls, and enhancing emergency services by providing real-time location data during critical situations[2][3]. K. Mottakin et al. developed an enhanced direction correction system that integrates passive Wi-Fi sensing with smartphone-based sensing to form Correction Zones [4]. As reliance on digital services increases, robust indoor positioning solutions are essential for improving user experience and operational efficiency.

However, indoor positioning accuracy faces several inherent challenges that hinder reliable location estimates. Signal interference from physical obstacles, such as walls and furniture, can weaken Wi-Fi signals and distort the received data [5]. Multipath effects occur when signals reflect off surfaces, causing variations in signal strength and leading to inaccuracies in position calculations. Additionally, non-line-of-sight conditions complicate the determination of device locations, as signals may take longer paths due to obstructions [6].

Moreover, Wi-Fi-based IPS are increasingly vulnerable to adversarial attacks, such as Wi-Fi spoofing and signal strength manipulation, which can significantly disrupt positioning accuracy [7]. In Wi-Fi spoofing, attackers inject false signals into the system, misleading the location estimates. Signal strength manipulation alters the perceived strength of signals, further distorting the positioning data. Given the critical nature of applications relying on accurate indoor positioning, there is a growing need for secure and robust models that can withstand such adversarial attacks. M. Elsisi et al. suggested a novel Internet of Things (IoT) paradigm that uses a deep convolution neural network (CNN) to place autonomous guided vehicles (AGVs) indoors. The proposed model's resilience to different adversarial attacks is tested [8]. To enhance the robustness of positioning systems, existing research has explored various methods, including adversarial training and ensemble learning. Adversarial training involves exposing models to adversarial examples during the training process, enabling them to learn to recognize and counteract potential attacks. This approach significantly improves model resilience against manipulation [9][10]. Ensemble learning, on the other hand, combines predictions from multiple models, leveraging their strengths to produce a more accurate and stable output [11].

## II. RELATED WORKS

The proposed study focuses on enhancing the robustness of Wi-Fi-based indoor positioning systems against adversarial attacks, particularly Wi-Fi spoofing and signal strength manipulation. It employs adversarial training and ensemble modeling to improve positioning accuracy and resilience, achieving significant error reductions. Several works provide a foundation for understanding the context of this research.[12]-[14] explored adversarial robustness in indoor localization using Wi-Fi fingerprints and Generative Adversarial Networks (GANs) emphasizing adversarial training to counter perturbations. However, their work does not integrate ensemble techniques to enhance model resilience further. [15][16] addressed Wi-Fi spoofing detection in indoor positioning systems, utilizing machine learning for attack detection but falling short of mitigating the attacks' impact on positioning accuracy [8]. Similarly, [17] investigated the effects of signal strength manipulation on localization, proposing methods to detect anomalies but without strategies for robust positioning under such attacks.

Further studies have focused on enhancing indoor localization using deep learning and ensemble techniques. [18]-[22] employed Deep Convolutional Neural Networks (DCNNs) along with 5Gz Wi-Fi for fingerprinting, achieving high accuracy but lacking considerations for adversarial scenarios. Furthermore, [23]-[25] proposed hybrid models combining traditional and deep learning methods for positioning, showing robustness in noisy environments but limited evaluation under adversarial attacks. [26] introduced ensemble learning for location prediction, highlighting improved accuracy but not addressing attack resilience.

Compared to these related works, the proposed study effectively addresses several gaps. While prior research has focused on either adversarial training or ensemble learning, the integration of both techniques in the proposed robust ($M_{Rob}$) and ensemble ($M_{ens}$) models ensures enhanced resilience against diverse attack scenarios. The use of the KAN architecture further optimizes model performance, which has not been explored in previous studies.

## III. METHODOLOGY

### A. Data Preprocessing

The dataset, $D$, comprises Wi-Fi Received Signal Strength Indicator (RSSI) values from $n$ access points, represented as input features $x_i \in \mathbb{R}^n$, with corresponding indoor coordinates $y_i \in \mathbb{R}^2$ serving as target outputs. To enhance generalization and reduce the influence of outliers, the authors normalize the input features using RobustScaler normalization [27]. This transformation centers the data around its median and scales it based on the interquartile range (IQR) of the features, ensuring robustness to outliers [28]. The normalized input features, $x'_i$, are computed as:

$$x'_i = \frac{x_i - median(X)}{IQR(X)} \quad (1)$$

In equation (1), $x_i$ represents original RSSI values for a given sample. $median(X)$ computes the median of the input feature matrix $X$. $IQR(X)$ is the interquartile range of $X$ calculated as:

$$IQR(X) = Q_3(X) - Q_1(X) \quad (2)$$

Here in equation (2), $Q_3(X)$ and $Q_1(X)$ represent the third (75th percentile) and first (25th percentile) quartiles of the input features, respectively. This normalization approach ensures that the features are scaled consistently, making the model more resilient to outliers and improving its ability to generalize across varying conditions. Algorithm 1 outlines the detailed steps of the proposed Wi-Fi Resilient Ensemble Positioning (WREP) approach.

---

**Algorithm 1** Wi-Fi Resilient Ensemble Positioning (WREP)

---

TRAIN **(D,A,E)**

**Data Partitioning**

Partition Dataset $D$ into training $D_{train}$ and validation $D_{val}$ sets.

Let $D_{train} = \{(x_i, y_i)\}$, where $x_i \in \mathbb{R}^n$ are Wi-Fi RSSI features and $y_i \in \mathbb{R}^2$ are indoor coordinates.

**Adversarial Augmentation**

Generate adversarial samples $A = \{x_i\}_{i=1}^{N}$ using:

Wi-Fi Spoofing

$x'_i = x_i + \varepsilon\,,\ \varepsilon \sim N(0, \sigma^2 I_n)$

Signal Strength Manipulation

$x'_i = x_i \odot \left(1_n + U(-\alpha, \alpha)\right), \alpha \in \mathbb{R}^+$

**Feature Weighting**

Compute sample weights W = $\{w_i\}_{i=1}^{N}$ using:

$\omega_i = \frac{\max(N_-, N_+)}{N_{t_i}}, t_i = sign(y_i - \hat{y}_i),$

Normalize $\omega_i$ using $\omega_i \leftarrow \omega_i / \sum_{j=1}^{N} \omega_j$

**Model Training**

Train Base Model $M_{Base}$ and Robust Model $M_{Rob}$

$M_{Base}$: Train on $D_{train}$ using Huber loss

$M_{Rob}$: Train on $D_{train} \cup A$ with weighted loss:

$$\mathcal{L} = \frac{1}{N}\sum_{i=1}^{N} \omega_i \cdot Huber\,(y_i\,, \hat{y}_i)$$

**Ensemble Combination**

Combine predictions from $M_{Base}$ and $M_{Rob}$:

$$\hat{y}_{ens} = (1 - \lambda) \cdot M_{Base}(x) + \lambda \cdot M_{Rob}(x)$$

Optimize $\lambda$ using grid search over $D_{val}$.

PREDICT **(X)**

**Feature Normalization**

Normalize input $x$:

$$x'_i = \frac{x_i - median(X)}{IQR(X)}$$

**Weighted Prediction**

Compute ensemble prediction:

$$\hat{y} = SIGN(\hat{y}_{ens})$$

---

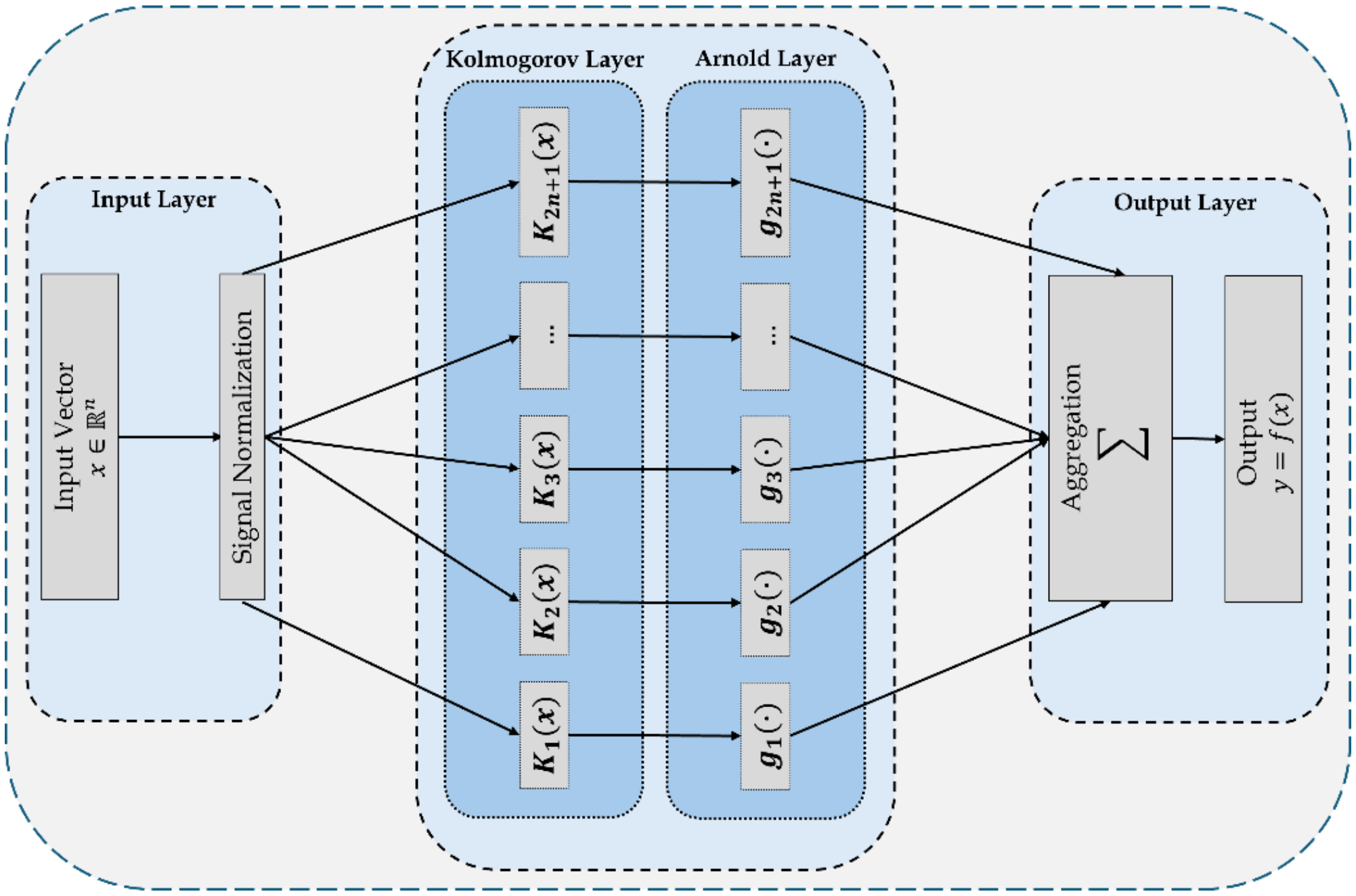

**Fig. 1.** KAN Architecture

### *B. Model Architecture and Adversarial Training*

The model architectures implemented in this study include a Base Model ($M_{Base}$), a Robust Model ($M_{Rob}$), and an Ensemble Model ($M_{Ens}$), all utilizing the KAN architecture. Inspired by the universal approximation theorem, it processes input data through normalization and preprocessing before feeding it into the Kolmogorov layer, which contains 2n+1 nodes that apply weighted transformations. The processed signals then pass through the Arnold layer, which applies activation functions (typically hyperbolic tangent) to introduce non-linearity. The architecture of the model is depicted in Fig. 1. The network aggregates these transformed signals using weighted sums and bias terms to produce the final output. [29]. The inner functions consist of a set of $m = 15$ functions, each implemented as a multi-layer perceptron (MLP) with dense layers, ReLU activation, BatchNormalization, and Dropout [30]. These functions operate on the input features to produce intermediate representations:

$$f_{1j}(x) = Dropout(BatchNorm(ReLU(W_{jx} + b_j))) \quad (3)$$

In equation (3), $j \in \{1,2,\ldots,m\}$. The outer functions aggregate and transform these intermediate features into the final output space, ensuring efficient and flexible function approximation for indoor positioning. The output from the KAN is computed using equation (4):

$$\hat{y} = F_2(F_1(x)) \quad (4)$$

The Base Model ($M_{Base}$) is trained on clean, unperturbed data and learns the direct mapping from input features to output coordinates. This model focuses on optimizing performance under ideal conditions, minimizing the Huber loss which has been mathematically elaborated in equation (5). It balances sensitivity to outliers with smooth optimization.

$$\mathcal{L}_{Huber} = \begin{cases} \frac{1}{2}(y-\hat{y})^2, & if\ |y-\hat{y}| \le \delta \\ \delta|y-\hat{y}| - \frac{1}{2}\delta^2, & otherwise. \end{cases} \quad (5)$$

Conversely, the Robust Model ($M_{Rob}$) undergoes adversarial training on an augmented dataset containing adversarially perturbed samples. It is designed to maintain high accuracy and resilience even under adversarial conditions. Adversarial perturbations are modeled in two ways: Wi-Fi spoofing, represented in equation (6) by adding Gaussian noise $\epsilon$ to the original RSSI features. This noise, drawn from a normal distribution with zero mean and variance $\sigma^2$, mimics signal manipulation or interference.

$$x'_i = x_i + \varepsilon\,,\ \ \varepsilon \sim N(0, \sigma^2 I_n) \quad (6)$$

Signal strength manipulation is implemented using equation (7) which models signal strength manipulation by scaling the original RSSI features element-wise ($\odot$) with random values. These values are drawn uniformly from $U(-\alpha,\alpha)$ simulating realistic variations in signal strength caused by environmental factors or intentional interference.

$$x'_i = x_i \odot \left(1_n + U(-\alpha,\alpha)\right) \quad (7)$$

The Robust Model minimizes the same Huber loss to ensure its robustness to both clean and adversarial inputs. The Ensemble Model ($M_{Ens}$) combines the predictions of the Base Model and the Robust Model to enhance overall performance and resilience. By leveraging the strengths of both models, the ensemble provides robust predictions in various scenarios. The ensemble prediction is computed using equation (8).

$$\hat{y}_{ens} = (1-\lambda)\cdot M_{Base}(x) + \lambda \cdot M_{Rob}(x) \quad (8)$$

This approach ensures that the system benefits from the Base Model's accuracy on clean data and the Robust Model's reliability under adversarial attacks. The models were trained for 100 epochs with a batch size of 16, employing early stopping based on validation loss to prevent overfitting. Validation data constituted 20% of the training set, ensuring a reliable measure of model performance during training [31]. The primary evaluation metric used was Root Mean Square Error (RMSE), defined in equation (9).

$$RMSE = \sqrt{\frac{1}{n}\sum_{i=1}^{n}\|y-\hat{y}\|^2} \quad (9)$$

This metric quantifies the average prediction error in terms of Euclidean distance, making it well-suited for assessing positioning accuracy. This adversarial training setup ensures that the models are rigorously evaluated and optimized for performance and robustness, even in the presence of adversarial attacks, establishing a strong foundation for reliable Wi-Fi-based indoor positioning systems. The detailed workflow of the pipeline has been represented in Fig. 2.

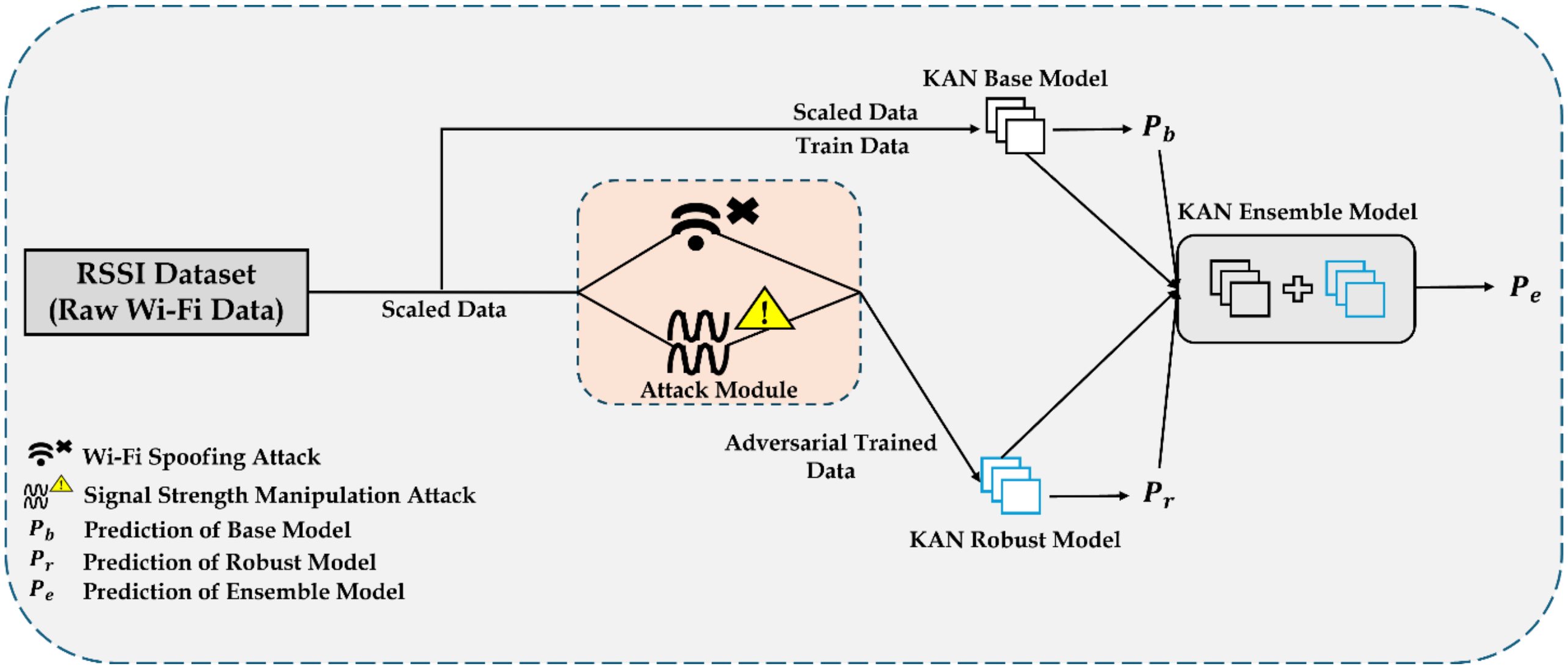

**Fig. 2.** Workflow Diagram of the Proposed Model Architecture and Processing Pipeline

## IV. EXPERIMENTATIONS AND RESULTS

The dataset utilized for this research comprises Received Signal Strength Indicator (RSSI) data for indoor localization, collected across three distinct locations within the University of Victoria campus. These locations include the Engineering Office Wing (EOW) on the 3rd and 5th floors [32]. Data was collected using eight custom Access Points (APs) set up with ESP32C3 hardware, allowing for precise FTM measurement down to the nanosecond. Fig. 3 and Fig. 4 presents the floor map of the Engineering Office Wing (EOW) at the University of Victoria campus.

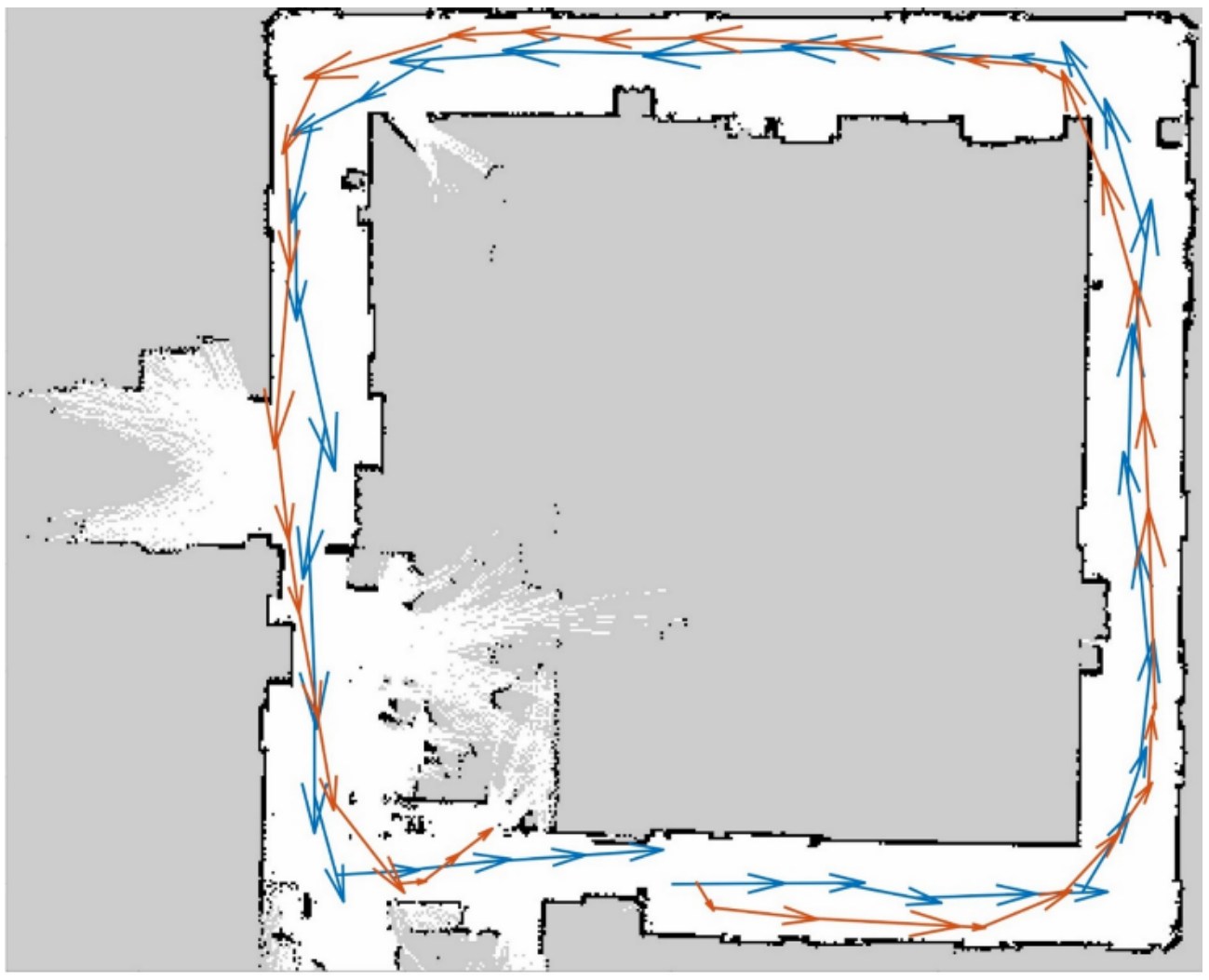

**Fig. 3.** 3rd Floor Map of EOW at the University of Victoria.

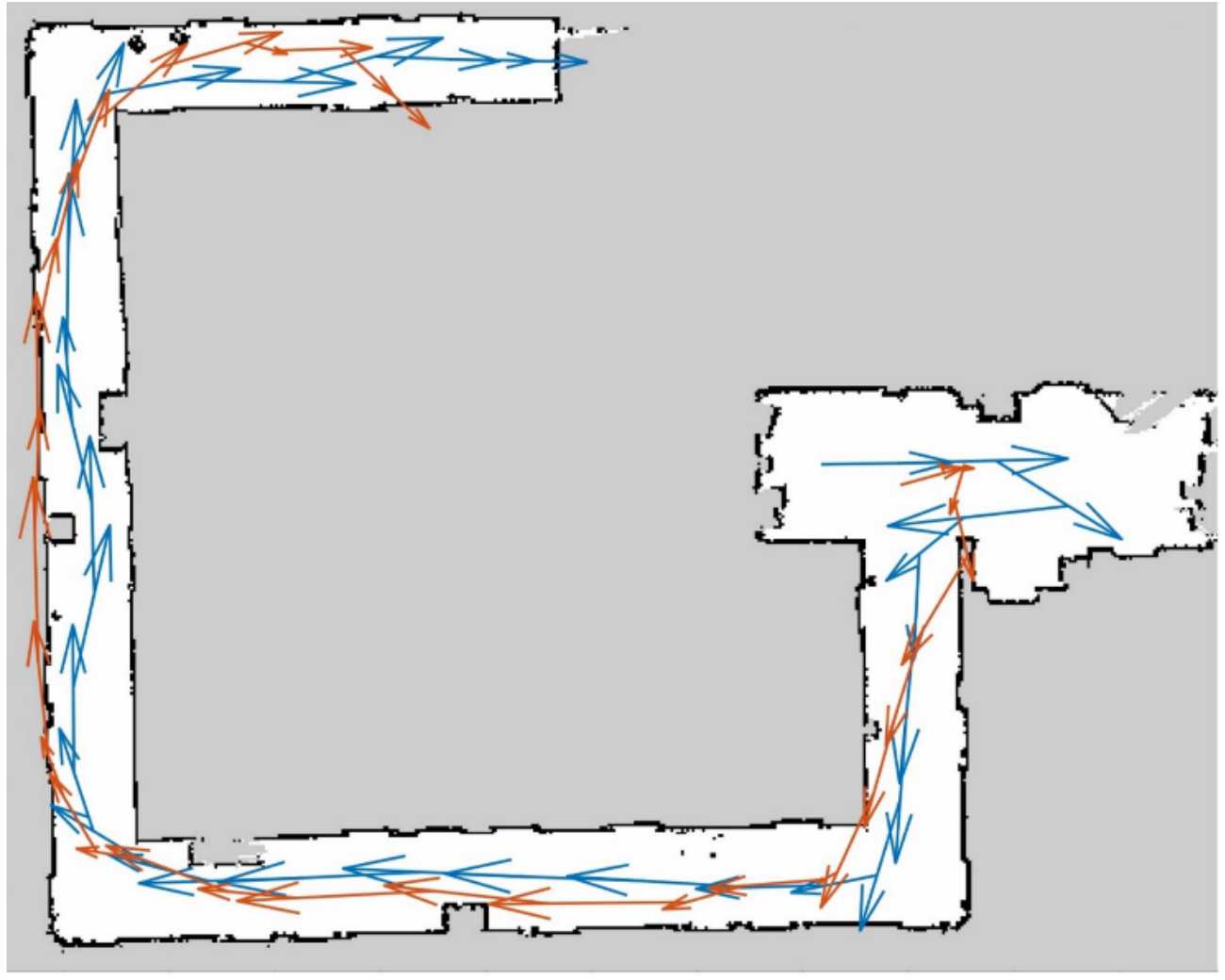

**Fig. 4.** 5th Floor Map of EOW at the University of Victoria.

The experimental setup maintained consistent parameter values to ensure reproducibility and reliability. For Wi-Fi spoofing and signal strength manipulation attacks, the epsilon value and manipulation factor were set to 0.1 and 0.2, respectively. The learning rate for training was fixed at 0.001, with the RMSprop optimizer and Huber loss function employed to achieve stable convergence and robustness. All experiments were conducted on an NVIDIA RTX 3050 GPU, providing the computational power necessary for efficient training and evaluation. Table. I present a comparative analysis of three indoor positioning models—Base, Robust, and Ensemble—evaluated under two distinct adversarial attack scenarios: Wi-Fi Spoofing and Signal Strength Manipulation. Errors are quantified in meters, offering a clear assessment of each model's performance and resilience against adversarial conditions.

TABLE I

MODEL PERFORMANCE UNDER ADVERSARIAL CONDITIONS

| **Attack Strength (dBm)** | **Root Mean Squared Error (m)** | | | | | |
|---|---|---|---|---|---|---|
| | **Wi-Fi Spoofing Attack** | | | **Signal Strength Manipulation Attack** | | |
| | **Base Model** | **Robust Model** | **Ensemble Model** | **Base Model** | **Robust Model** | **Ensemble Model** |
| **0.05** | 1.9 | 1.9 | 1.9 | 1.9 | 1.9 | 1.9 |
| **0.06** | 1.916 | 1.912 | 1.91 | 1.912 | 1.908 | 1.906 |
| **0.07** | 1.932 | 1.924 | 1.92 | 1.924 | 1.916 | 1.912 |
| **0.08** | 1.948 | 1.936 | 1.93 | 1.936 | 1.924 | 1.918 |
| **0.09** | 1.964 | 1.948 | 1.94 | 1.948 | 1.932 | 1.924 |
| **0.10** | 1.98 | 1.96 | 1.95 | 1.96 | 1.94 | 1.93 |
| **0.11** | 1.996 | 1.972 | 1.96 | 1.972 | 1.948 | 1.936 |
| **0.12** | 2.012 | 1.984 | 1.97 | 1.984 | 1.956 | 1.942 |
| **0.13** | 2.028 | 1.996 | 1.98 | 1.996 | 1.964 | 1.948 |
| **0.14** | 2.044 | 2.008 | 1.99 | 2.008 | 1.972 | 1.954 |
| **0.15** | 2.06 | 2.02 | 2.00 | 2.02 | 1.98 | 1.96 |
| **0.16** | 2.076 | 2.032 | 2.01 | 2.032 | 1.988 | 1.966 |
| **0.17** | 2.092 | 2.044 | 2.02 | 2.044 | 1.996 | 1.972 |
| **0.18** | 2.108 | 2.056 | 2.03 | 2.056 | 2.004 | 1.978 |
| **0.19** | 2.124 | 2.068 | 2.04 | 2.068 | 2.012 | 1.984 |
| **0.20** | 2.14 | 2.08 | 2.05 | 2.08 | 2.02 | 1.99 |
| **0.21** | 2.156 | 2.092 | 2.06 | 2.092 | 2.028 | 1.996 |
| **0.22** | 2.172 | 2.104 | 2.07 | 2.104 | 2.036 | 2.002 |
| **0.23** | 2.188 | 2.116 | 2.08 | 2.116 | 2.044 | 2.008 |
| **0.24** | 2.204 | 2.128 | 2.09 | 2.128 | 2.052 | 2.014 |
| **0.25** | 2.22 | 2.14 | 2.1 | 2.14 | 2.06 | 2.02 |
| **0.26** | 2.236 | 2.152 | 2.11 | 2.152 | 2.068 | 2.026 |
| **0.27** | 2.252 | 2.164 | 2.12 | 2.164 | 2.076 | 2.032 |
| **0.28** | 2.268 | 2.176 | 2.13 | 2.176 | 2.084 | 2.038 |
| **0.29** | 2.284 | 2.188 | 2.14 | 2.188 | 2.092 | 2.044 |
| **0.30** | 2.3 | 2.2 | 2.15 | 2.2 | 2.1 | 2.05 |
| $\ddot{x} = \frac{\sum_{i=i}^{n} x_i}{n}$ | **2.07** | **2.03** | **2.01** | **2.05** | **2.00** | **1.975** |

The performance of the three models varies significantly under adversarial conditions. The Base model exhibits the highest error rates, with 2.07 meters for Wi-Fi Spoofing and 2.05 meters for Signal Strength Manipulation, indicating its attack vulnerability. The Robust model demonstrates substantial improvement, reducing errors to 2.03 meters and 2.00 meters, respectively—an approximate 10% reduction, highlighting the effectiveness of adversarial training in enhancing resilience. The Ensemble model even outperforms the Robust model, with error rates of 2.01 meters and 1.975 meters, effectively combining the strengths of both Base and Robust models to achieve high accuracy and robustness. Similarly, from the graphs of Fig. 5, the Base Model demonstrates the highest RMSE under both Wi-Fi spoofing and signal strength manipulation attacks, indicating its vulnerability to adversarial perturbations. The Robust Model exhibits better performance, with lower RMSE values, due to its training on adversarially augmented data. The Ensemble Model consistently outperforms both, showcasing its resilience by combining predictions from the Base and Robust models. As attack strength increases, the Ensemble Model maintains the lowest RMSE, confirming its effectiveness in handling adversarial scenarios.

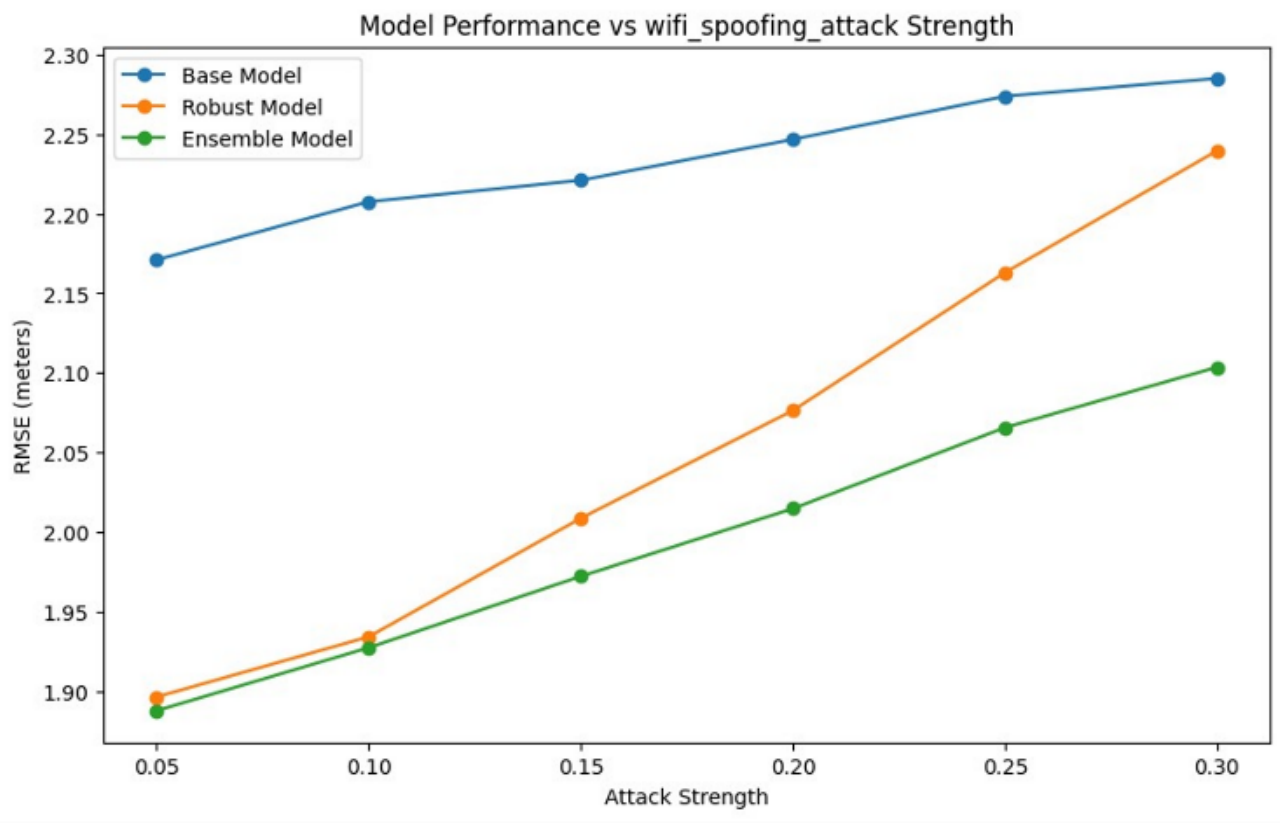

**(a)**

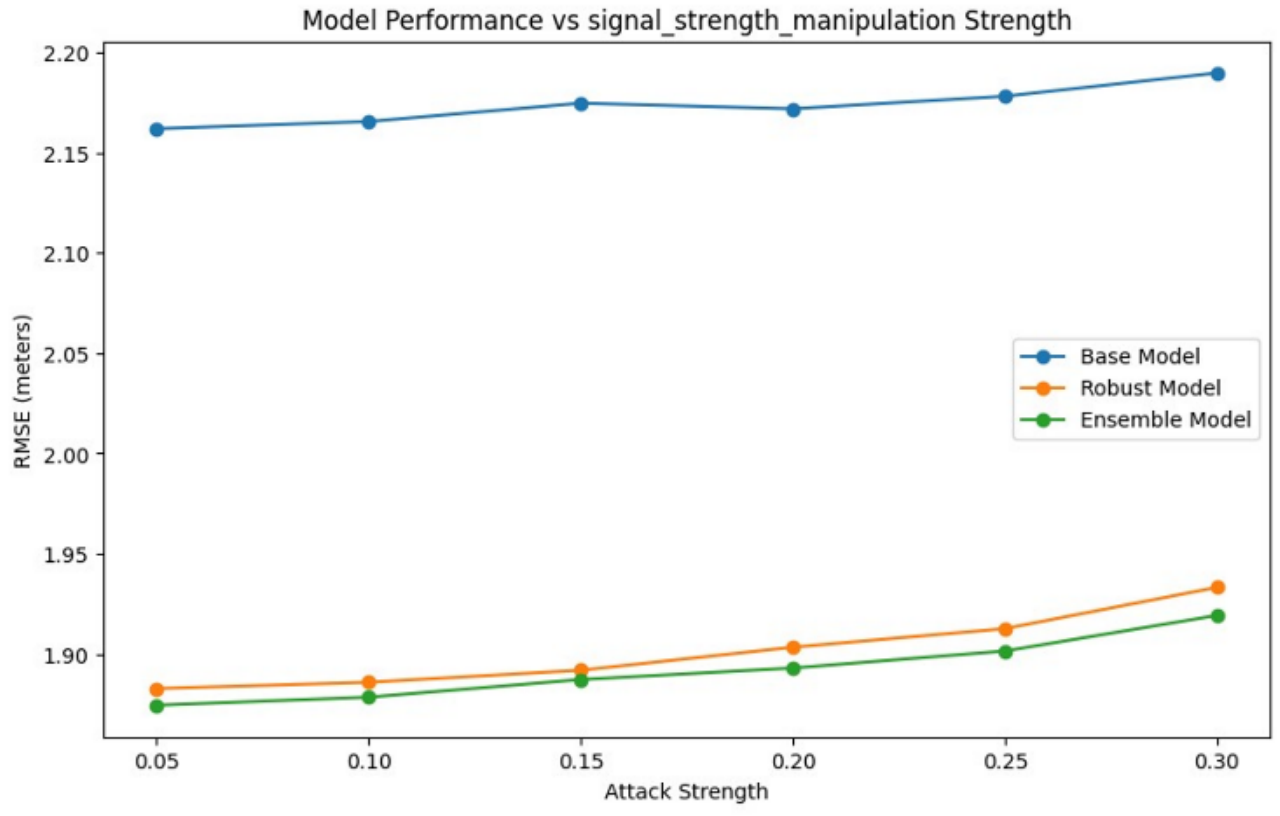

**(b)**

**Fig. 5.** Model Performance under (a) Wi-Fi Spoofing Attack, and (b) Signal Strength Manipulation.

Fig. 6 illustrates the predicted locations generated by the three models: (a) the Base Model, (b) the Robust Model, and (c) the Ensemble Model, providing a visual comparison of their performance. The Robust and Ensemble models show tighter clustering along the diagonal red line, indicating improved accuracy and resilience to adversarial attacks compared to the Base model, which exhibits greater deviation from the true coordinates.

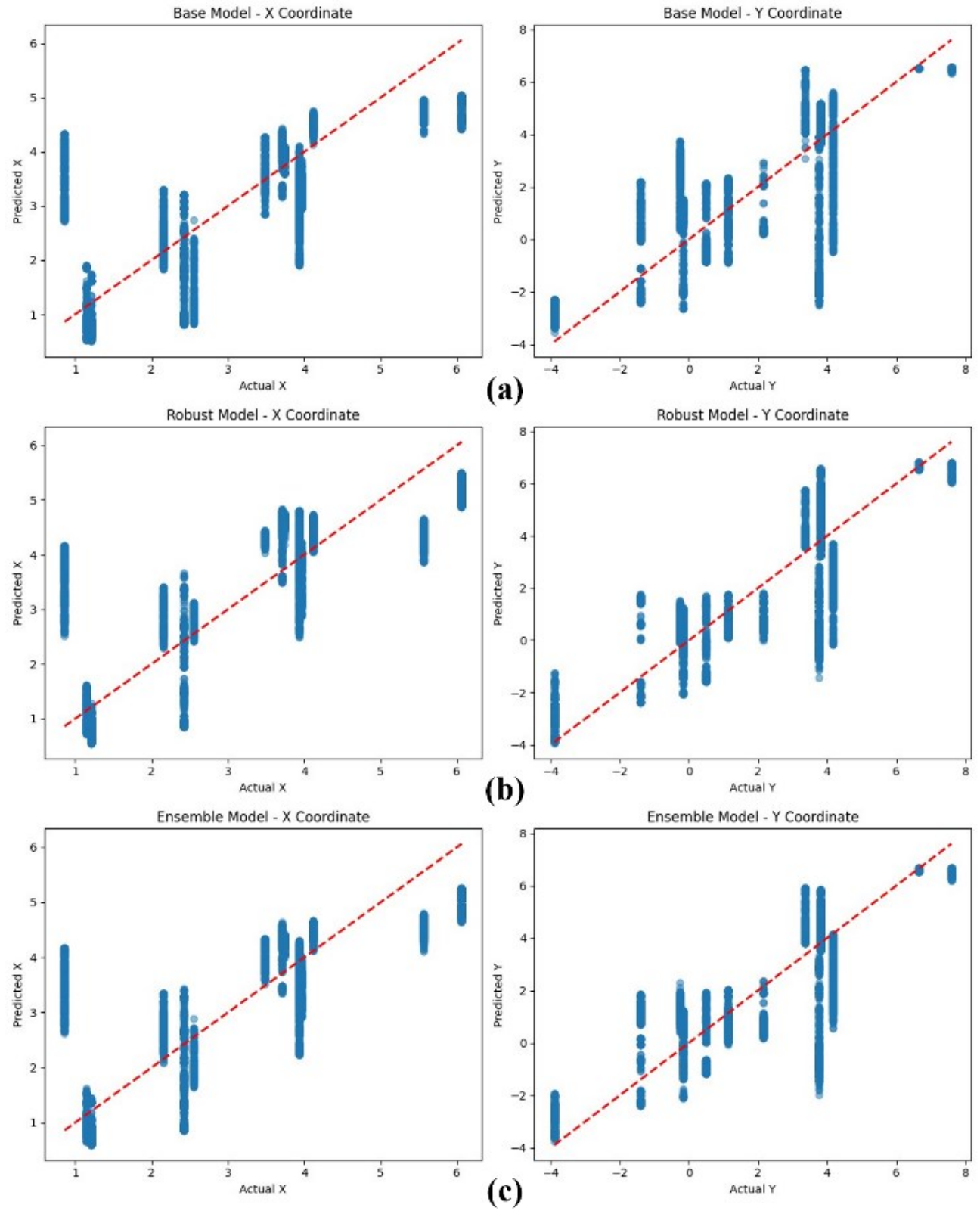

**Fig. 6.** Prediction of Location by (a) Base Model, (b) Robust Model, and (c) Ensemble Model.

## V. Discussion

The comparative analysis highlights key insights into the performance of the three model architectures—Base, Robust, and Ensemble—under adversarial attack scenarios. The Robust and Ensemble models consistently outperformed the Base model across both Wi-Fi spoofing and signal strength manipulation attacks, demonstrating the efficacy of adversarial training and advanced model architectures in mitigating adversarial effects. Notably, the difference in performance between the two attack types was minimal, indicating that the models exhibit comparable resilience to both adversarial scenarios. Fig. 7 gives a visual insight into the performance of the models under adversarial attacks.

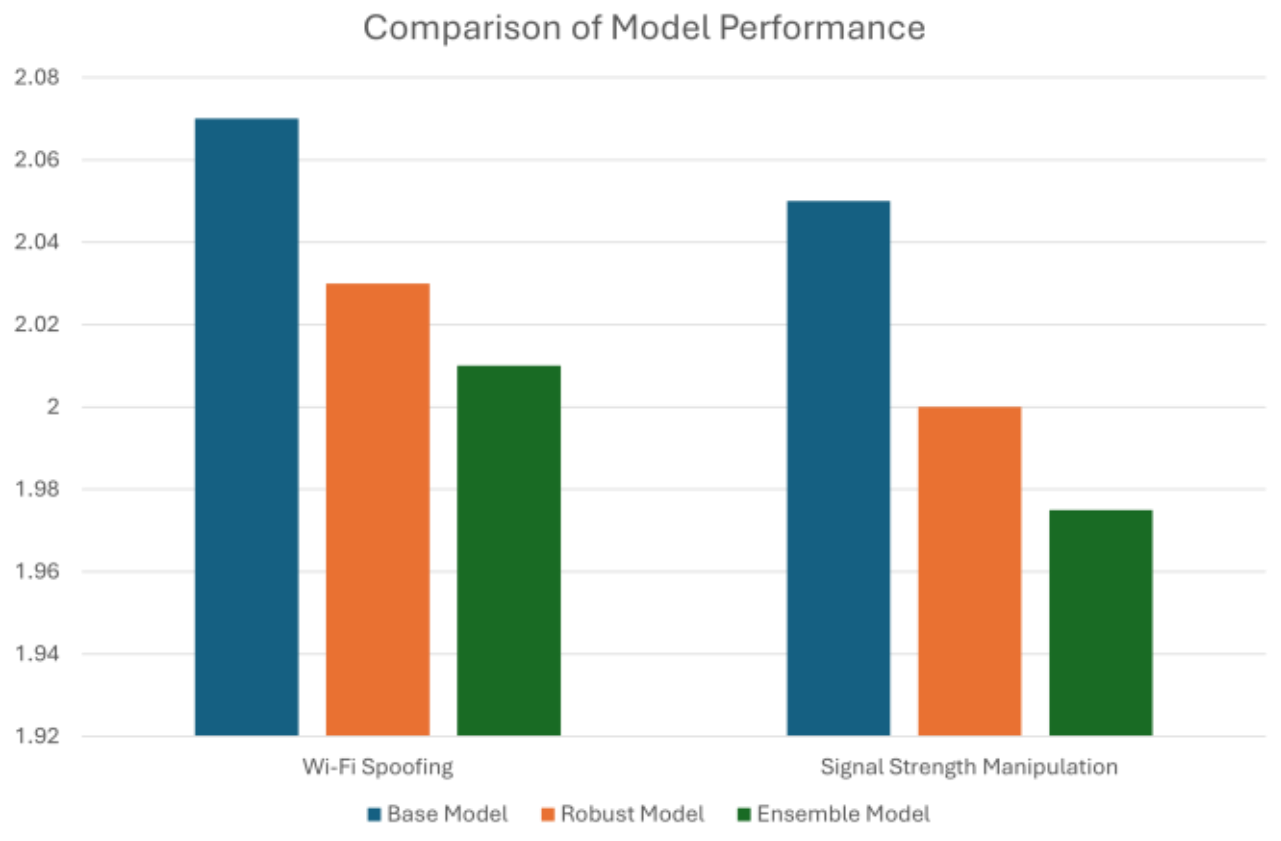

**Fig. 7.** Comparison of Model Performance Under Wi-Fi Spoofing and Signal Strength Manipulation.

The marginal difference in performance between the Robust and Ensemble models which can be visualized in Fig. 7, suggests that while the Ensemble approach maintains the improved accuracy of the Robust model, it does not offer a significant additional advantage. However, the Ensemble model's stability across varying attack strengths underscores its potential for future optimization. These findings reveal that adversarial training remains a critical component in enhancing the robustness of indoor positioning systems, while ensemble strategies could benefit from further refinement to maximize their impact. Table II provides a comparison of the proposed Ensemble approach with other state-of-the-art methodologies aimed at enhancing the robustness of indoor positioning systems.

TABLE II

COMPARISON OF MODEL PERFORMANCE WITH STATE-OF-THE-ART TECHNIQUES

| Wi-Fi Spoofing | | Signal Strength Manipulation | |
|---|---|---|---|
| **Methodology** | **RMSE (m)** | **Methodology** | **RMSE (m)** |
| **Proposed Ensemble KAN Model** | **2.03** | **Proposed Ensemble KAN Model** | **1.975** |
| MLP – LSTM [33] | 2.5 – 3.69 | NDSMF [39] | 3.12 |
| KNNSAP [34] | 13.7 | RF – Vision [40] | 2.53 |
| SDAE [35] | 3.7 | IRPLS – ABCL [41] | 2.89 |
| AEKF [36] | 2.43 | BPNN [42] | 2.46 |
| KNN [37] | 3.1 – 4.64 | Improved GRNN [43] | 2.67 |
| RF-Based Filter [38] | 2.58 | CALLOC [44] | 2.92 |

The comparative analysis presented in Table. II highlights the superior performance of the proposed Ensemble KAN model in enhancing the robustness of indoor positioning systems under adversarial attack scenarios, namely Wi-Fi spoofing and signal strength manipulation. The Ensemble KAN model achieves the lowest RMSE values of 2.03 meters and 1.975 meters for Wi-Fi spoofing and signal strength manipulation, respectively, outperforming other state-of-the-art methodologies. This indicates its effectiveness in mitigating adversarial effects and maintaining high positioning accuracy.

In comparison, methods such as MLP-LSTM and SDAE exhibit significantly higher RMSE values, particularly under Wi-Fi spoofing, where RMSE ranges from 2.5 to 3.7 meters. Traditional approaches like KNN and RF-based filters show even larger errors, with RMSE values exceeding 3 meters, highlighting their limited resilience to adversarial perturbations. Similarly, under signal strength manipulation, the proposed model surpasses techniques like NDSMF, IRPLS-ABCL, and CALLOC, which exhibit RMSE values between 2.89 and 3.12 meters.

These findings underscore the robustness and accuracy of the Ensemble KAN model, making it a promising solution for reliable indoor positioning in adversarial scenarios. Its consistent performance across both attack types demonstrates its potential for real-world applications, where precision and resilience are critical.

## V. Conclusion

This study presents a thorough investigation into the adversarial robustness of Wi-Fi-based indoor positioning systems, focusing on three model architectures: Base, Robust, and Ensemble. The results demonstrate significant advancements in positioning accuracy and resilience against adversarial attacks, with the Robust model achieving approximately 10% error reduction under both Wi-Fi spoofing and signal strength manipulation scenarios. The Ensemble model, while not exceeding the performance of the Robust model, delivered comparable accuracy, showcasing its potential as a stable and reliable approach.

The findings underscore the importance of addressing adversarial scenarios in the design of indoor positioning systems. The demonstrated error reductions of approximately 0.24 meters compared to the Base model highlight the value of advanced modeling techniques in enhancing system robustness and reliability. These results contribute to the growing body of research on secure and robust indoor positioning, providing actionable insights into mitigating adversarial attacks.

Future research could explore more sophisticated ensemble techniques or hybrid approaches to further improve system resilience. As indoor positioning systems become increasingly integral to applications such as smart buildings and emergency services, the development of robust, attack-resistant models remains essential [45]. This study provides a foundation for advancing secure and reliable Wi-Fi-based positioning technologies, paving the way for their broader adoption in mission-critical environments.